\documentclass[runningheads,a4paper]{llncs}
\pdfoutput=1

\usepackage{amssymb}
\usepackage{graphicx}
\usepackage{booktabs}
\usepackage{array}
\usepackage{multirow}
\usepackage{float}
\usepackage[dvipsnames,table]{xcolor}
\usepackage{url}
\usepackage{tikz}
\usepackage{caption}
\usepackage{subcaption}
\usepackage{tabularx}
\usepackage{pbox}
\usepackage{multirow}
\usepackage{lipsum}
\usepackage{url}

\urldef{\mailsa}\path|{alfred.hofmann, ursula.barth, ingrid.haas, frank.holzwarth,|
\urldef{\mailsb}\path|anna.kramer, leonie.kunz, christine.reiss, nicole.sator,|
\urldef{\mailsc}\path|erika.siebert-cole, peter.strasser, lncs}@springer.com|    
\newcommand{\keywords}[1]{\par\addvspace\baselineskip
\noindent\keywordname\enspace\ignorespaces#1}
\newcolumntype{C}[1]{>{\centering\arraybackslash}p{#1}}
\newcolumntype{L}[1]{>{\arraybackslash}p{#1}}

\begin{document}

\mainmatter  

\title{A Novel Deep Learning Architecture for Testis Histology Image Classification}

\titlerunning{A Novel DL Architecture for Testis Histology Image Classification}

%
%
\author{Chia-Yu Kao and Leonard McMillan}
\authorrunning{Chia-Yu Kao and Leonard McMillan}

\institute{Department of Computer Science, University of North Carolina, Chapel Hill, NC}

%
%

\toctitle{A Novel Deep Learning Architecture for Testis Histology Image Classification}
\tocauthor{$Chia-Yu Kao and Leonard McMillan}
\maketitle

\begin{abstract}
Unlike other histology analysis, classification of tubule status in testis histology is very challenging due to their high similarity of texture and shape. Traditional deep learning networks have difficulties to capture nuance details among different tubule categories. In this paper, we propose a novel deep learning architecture for feature learning, image classification, and image reconstruction. It is based on stacked auto-encoders with an additional layer, called a hyperlayer, which is created to capture features of an image at different layers in the network. This addition effectively combines features at different scales and thus provides a more complete profile for further classification. Evaluation is performed on a set of 10,542 tubule image patches. We demonstrate our approach with two experiments on two different subsets of the dataset. The results show that the features learned from our architecture achieve more than 98\% accuracy and represent an improvement over traditional deep network architectures.
\end{abstract}

\section{Introduction}
To evaluate male fertility, sperm concentration, morphology and motility are common parameters measured in a semen analysis \cite{fernando}. However, a far richer and complementary source of information related to male fertility can be found via direct examination of histology images of testicle cross sections. Examining histology is only practical in model organisms, such as the laboratory mouse, but such examinations may eventually help to unravel the underlying bimolecular causes of infertility. A testis histology image can be segmented into a units known as seminiferous tubules. The visual state of each tubule is indicative of a many aspects of spermatogenesis, including the stage of meiosis \cite{Oakberg} and structural abnormalities. Traditionally, classification of tubule status is performed manually by pathologists. This process is time-consuming, labor-intensive, and prone to subjective interpretation \cite{derde}. Moreover, the scoring of samples is often inconsistent between pathologists, and even scores from the same pathologist are variable over time and dependent on local context. In this paper we examined whether machine-learning methods might be used to address the above problems and, meanwhile, to increase histological diagnostic efficiency and consistency.


\begin{table}
\caption{Examples of tubule histology image in different stages.}
\centering
\begin{tabular}{C{2cm} C{2cm} C{2cm} C{2cm} C{2cm}}
\includegraphics[width=0.1\textwidth]{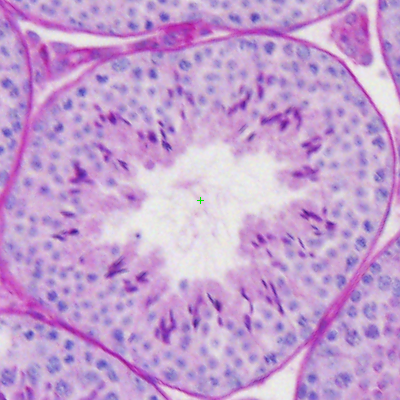}	&	\includegraphics[width=0.1\textwidth]{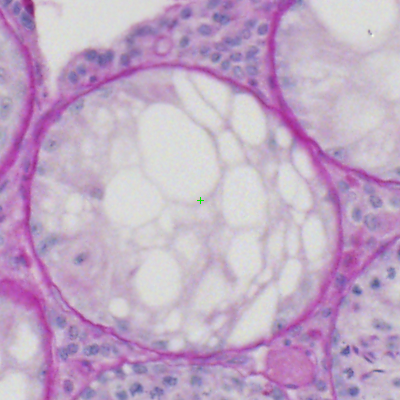}	&	\includegraphics[width=0.1\textwidth]{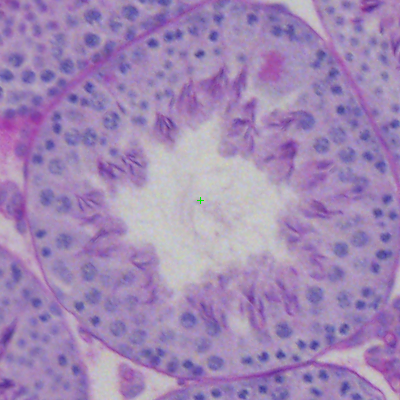}	&	\includegraphics[width=0.1\textwidth]{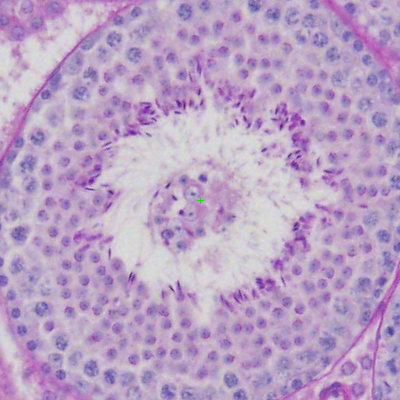}	&	\includegraphics[width=0.1\textwidth]{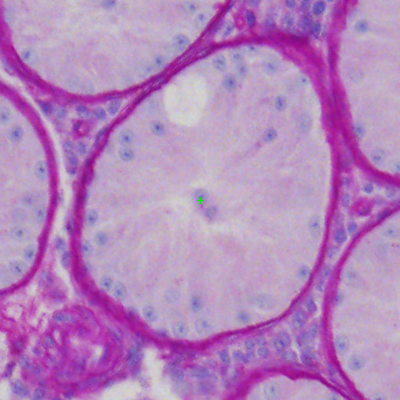}\\
Normal	&	Vacuoles	&	\parbox{2cm}{\centering Abnormal \\germ cells}	&	\parbox{2cm}{\centering Germ cells \\in lumen}	&	\parbox{2cm}{\centering Germ cell \\ loss}\\[-1em]
\end{tabular}      
\label{table:tubule}
\end{table}

\vspace*{-2em}
The paper presents a learning features, image reconstruction, and automatic classification of testis histology images. Unlike other histology image analysis to predict cell types, we classify the tubules in different states (a combination of spermatogenesis stage and abnormality) where most of the tubules have very similar texture and shape. Because of the similarity, it makes our classification problem more challenging than others. Table \ref{table:tubule} shows histology image samples stained with hematoxylin of seminiferous tubules exhibiting defects, including vacuoles, abnormal, loss of germ cells, and the unexpected sloughing of materials into the normally clear lumen of the tubule. In many aspects, a normal tubule at one stage of spermatogenesis might look very similar to an abnormal one. Because of this subtly, significant training is required before a human observer can differentiate between them. To solve this problem, we propose a novel deep learning architecture where an additional layer, called a ‘hyperlayer’, is created and combined with convolutional layers to capture high level semantic information and small scale image features. The proposed approach does not follow traditional convolutional network only taking the top layer for classification which may be too coarse to reveal the nuance among different stages of tubules.

The deep learning architecture composes multiple linear and non-linear transformations of the data and yields a more abstract and useful representation \cite{angel}. The attraction of this approach is that feature extraction is part of the learning process. Our goal is to build a learning architecture which can extract the appropriate features, visualize the features learned by the model, and classify the tubules based on the features learned. In our architecture, the learning network is able to capture features of an image from different levels of abstraction and scale and thus provide a more complete description for the classification process. Inspired by deep learning in image classification, the novel deep learning architecture combines convolutional auto-encoders, a hyperlayer, and a softmax classifier for tubule classification and visual representation. The main contributions of this work are presenting a novel deep learning architecture which achieves real-world performance comparable to human annotations while outperforming traditional stacked convolutional auto-encoders. In addition, our model provides functionality for visual representation of learned features. It could help make significant improvements on model training.

\section{Related Work}
In recent years, deep learning architectures have achieved unprecedented performance in different computer vision and pattern recognition tasks \cite{kriz,bengio}. Convolutional neural network (CNN) especially have had notable success in this area. Previous work using deep learning architectures for automatic classification and detection in biomedical images has primarily focused on detecting cell types or abnormal structures, such as malignant tumors, which are significantly different in appearance from the normal tissues, like breast cancer \cite{dan} or bacal-cell carcinoma \cite{angel}. However, our application is more challenging because we classify images with very similar texture and shape. 

The information presented in intermediate layers of a deep learning network tend to be more generalized in the top layer but less sensitive to semantics \cite{bharath}. The information of interest is distributed over all levels of the learning network and the features extracted for classification should learn from all of the layers instead of the previous layer only. Long et al. \cite{long} combine intermediate layers with the top layer in their convolutional segmentation architecture. Bharath \cite{bharath} define the "hypercolumn" at all layers of CNN for their object segmentation network. Tompson et al. \cite{jonathan} introduce a refinement model jointly trained in cascade with convolutional network for human joint location estimation.\\

\vspace*{-2em}
\section{Approach}
\begin{figure}[ht!]
\centering
\includegraphics[width=1\textwidth]{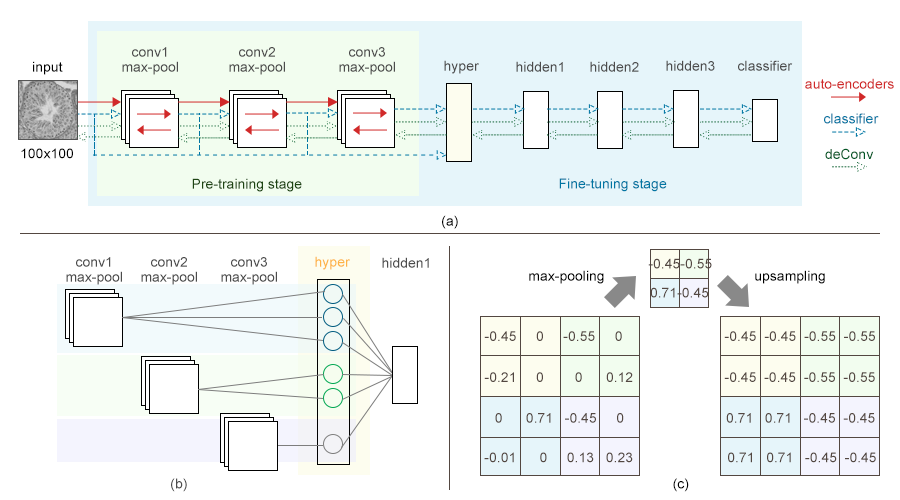}
\vspace*{-1em}
\caption{(a) Overview of our deep learning architecture, where the building blocks within green box are convolutional anto-encodes performed in the first stage of classification, and all building blocks within blue box compose a whole learning framework for the fine-tuning stage. Deconvolutional networks indicated in green dotted lines goes through whole framework to reconstruct the image from the top most layer. (b) Example of a hyper layer with weights 3, 2, and 1 on convolutional layer 1, 2, and 3, respectively. (c) Example of upsampling operation on pool size 2x2 in deconvolutional networks. Pool-sized neighborhoods are filled with the absolute max value within that area.}
\label{fig:overview}
\end{figure}

The proposed architecture for testis histology images learning, visualization of learning process, feature combination, and classification is based on deep learning network architecture depicted in Figure \ref{fig:overview}(a). We exploit stacked convolutional auto-encoders \cite{SCAE} for classification, which includes unsupervised pre-training stage (green block) and supervised fine-tuning stage (blue block). In pre-training stage, the architecture is trained via convolutional auto-encoders one layer at a time. Once all layers are pre-trained, the network goes through to the second fine-tuning stage, where we add our hyperlayer to capture features from all layers/scales from the first stage. As for visualization of learning process, we reconstruct the image with learned features via deconvolutional networks. Each module is corresponding to the layers in our architecture described in the following sections.\\

\noindent{}\textbf{Feature Learning via Stacked Convolutional Auto-Encoders:} The convolutional auto-encoders architecture is similar to auto-encoders, except the weights in convolutional neurons are shared to achieve spatial/position independence \cite{SCAE}. Several convolutional auto-encoders can be stacked to form a deep learning architecture. Each layer takes the max-pooling of the hidden outputs from the layer below as its input, and performs unsupervised pre-training. The pre-training process first takes the input $x$ and maps it to the $k$-$th$ feature map using the function $h^k=\sigma($x$*$W\textsuperscript{k}$+b^k)$ with parameters, W and $b$. $\sigma$ is an activation function, and * denotes the 2D convolution. Afterward, the feature maps are used to reconstruct the input by mapping back with the function $y=\sigma(\sum h^k*\tilde{W}^k+c)$ with parameters $\tilde{W}$ and $c$. Those parameters are optimized by minimizing the differences between input and the reconstructed representation. The learned features from convolutional auto-encoders are used to initialize a convolutional neural network with identical topology. Once the first stage unsupervised pre-training is done, the network goes through the convolutional neural network in the second fine-tuning stage as the blue dash lines shown in figure \ref{fig:overview}(a). \\

\vspace*{-3em}

\vspace*{-1em}
\noindent{}\textbf{Hyperlayer:} In the fine-tuning stage, unlike traditional stacked auto-encoders architecture, we add an intermediate layer, called hyperlayer, between convolutional layers and hidden layers to collect the features from different scales. A hyperlayer is a fully connected layer which concatenates extracted feature maps from every convolutional layer below and outputs the representative neurons of the corresponding layer to above layer. In addition, hyperlayer is allowed to specify the importance of every convolutional layer by configuring the weights for each of them. Figure \ref{fig:overview}(b) illustrates an example of the hyperlayer with weights 3, 2, and 1 on features collected from convolutional layer 1, 2, and 3, respectively. Combining all features across multiple layers provides a compact image representation with different scales of details for the classifier layer. \\

\noindent{}\textbf{Automatic Classification via Softmax Classifier:} Beyond the hyper layer, we construct several hidden layers to decrease the number of neurons in the framework and build a logistic regression classifier in the end of the architecture. The classifier is trained by minimizing the negative log-likelihood. The output of the classifier is the probability that the input $x$ is a member of a class $i$, and then we choose the class whose probability is maximal as our model's prediction.
\\

\noindent{}\textbf{Image Representation via Deconvolutional Networks:} We use convolutional auto-encoders to learn the features that can best reconstruct their input in a layer-wise fashion. Therefore, the reconstructed image reflects the features learned from the model in layers. To see the features learned from the whole framework including classification, we exploit deconvolutional networks \cite{deconv} to reconstruct the image from the classifier layer. The concept of the deconvolutional network is very similar to convolutional auto-encoders, except for the max-pooling part. In deconvolutional network, to reconstruct the image we have to upsample the results of max-pooling to form the hidden representation and then we are able to reconstruct the input, while auto-encoders can directly take the hidden representation to reconstruct the input. In our model, upsampling is operated by filling the max value in the pool-sized of neighborhoods. An example of upsampling is depicted in figure \ref{fig:overview}(c).

\section{Experimental Setup}
\textbf{Histology image dataset:} We evaluate our approach on a dataset including 6,588 tubule image patches of 100 x 100 extracted from testis histology images. The histology images are stained with hematoxylin and scanned with 20x magnification. The dataset contains five categories: normal seminiferous tubule, tubules with vacuoles, tubules with abnormal germ cells, tubules with germ cells in the lumen, and tubules with germ cell loss. Examples of tubules in these five categories are shown in table \ref{table:tubule}. The five categories are not mutually exclusive, and hence, it is not trivial to classify those tubule images. The images are manually annotated by an expert in male reproduction, indicating different combination of categories. To increase the statistical power, we replicate the abnormal tubules by rotating them in four different angles. We conducts two experiments on two different subsets of the dataset, each of them has 10,000 images in total. The first experiment takes the images from normal tubules (8,946) and tubules with vacuoles (1,054). The second one uses the subset of normal tubules (8,727), tubules with vacuoles, tubules with germ cell loss, and tubules with both vacuoles and germ cell loss (total 1,273 for the abnormal tubules). \\
	
\noindent{}\textbf{Building the classifier:} We partition our images into three subsets: training dataset, test dataset, and validation dataset. The datasets was randomly partitioned into 6 equal sized subsamples. Validation and testing datasets each used a partition, and the remaining 4 were used as training data. We train our approach on data in training dataset, and classify the tubules in test dataset. In order to choose the appropriate hyperparameters for our approach, we exhaustively perform parameter combinations exploration on data in validation dataset. In our classifier, we have three convolutional auto-encoders layers, one hyperlayer, and three fully-connected hidden layers. The combination of optimized hyperparameters is listed in table \ref{table:params}. We normalize all images between -1 and 1, and use hyperbolic tangent as our activation function, which maintains a consistent input range for all layers. Note that when we convolve the input, the surrounding missing part are padded with the input to keep the dimension of output image equivalent to the input.  \\

\noindent{}\textbf{Classification Performance and Comparison:} To evaluate our experiment results, we compare our approach against traditional CNN and standard stacked convolutional auto-encoders (which does not include the hyperlayer) with same hyperparameters. 

\vspace*{-2.5em}
\begin{table}\scriptsize
\centering
\caption{The hyperparameter settings for all layers in our learning architecture. Layer type: I - input, C - convolutional layer, MP - max-pooling layer, H - hyper layer, FC - fully-connected hidden layer, LG - logistic regression layer.\\}
\begin{tabular}{c C{1cm} C{5cm} C{2cm} C{2cm}}
\toprule[0.04cm]
Layer	&	Type	&	Maps and neurons	& Filter size	&	Weights\\
\hline
0		&	I	&	1M x 100x100N		&	-		&	- \\
1		&	C	&	50M x 50x50N		&	13x13	&	50 \\
2		&	MP	&	50M x 25x25N		& 	2x2		&	- \\
3		&	C	&	100M x 25x25N		&	9x9		&	100 \\
4		&	MP	&	100M x 13x13N		&	2x2		&	- \\
5		&	C	&	150M x 13x13N		&	3x3		&	150 \\
6		&	MP	&	150M x 7x7N		&	2x2		&	- \\
7		&	H	&	900N			&	-		&	- \\
8		&	FC	&	512N			&	-		&	- \\
9		&	FC	&	100N			&	-		&	- \\
10		&	FC	&	50N				&	-		&	- \\
11		&	LG	&	2N				&	-		&	- \\
\bottomrule[0.04cm]
\end{tabular}
\label{table:params}
\end{table}
\vspace*{-4.8em}
\begin{table}
\centering
\caption{Comparison of classification performance on the first dataset, where SCAE is Stacked Convolutional Auto-Encoders. We compare our method with traditional CNN and SCAE with same parameters. The best results are in bold typeface.\\}
\begin{tabular}{L{4cm} C{2cm} C{2cm} C{2cm} C{2cm}}
\toprule[0.04cm]
	&	Accuracy	&	Precision	&	Recall	&	Specificity \\
\hline
CNN		&	0.901	&	0.348	&	0.862	&	0.904\\
SCAE	&	0.973	&	0.981	&	0.964	&	0.982\\
Proposed methods	&	\textbf{0.986}	&	\textbf{0.987}	&	\textbf{0.984}	&	\textbf{0.988}	\\
\bottomrule[0.04cm]
\end{tabular}
\label{table:performance}
\end{table}
\vspace*{-4.8em}
\begin{table}[!h]
\centering
\caption{Comparison of classification performance on the second dataset. The hyperparameters are same in these three methods. The best results are in bold typeface.\\}
\begin{tabular}{L{3cm} C{3cm} C{3cm} C{3cm}}
\toprule[0.04cm]
	&	CNN	&	SCAE	&	Proposed methods\\
\hline
Error rate	&	5.269	&	3.533	&	\textbf{2.150}	\\
\bottomrule[0.04cm]
\end{tabular}
\label{table:performance2}
\end{table}

\vspace*{-3em}
\section{Results and Discussion}
\textbf{Automatic Classification Performance: }
Table \ref{table:performance} shows the classification performance results in terms of accuracy, precision, recall, and specificity on the first dataset. Table \ref{table:performance2} compares the error rate among traditional CNN, SCAE, and our approach on the second dataset. The results show that our approach outperforms the canonical methods on a testis dataset. The fact of our architecture produces better results suggests that the proposed architecture has captured and learned the important features. Including a hyperlayer preserves the details at the lower layer which are critical in our dataset, while the traditional CNN or standard stacked convolutional auto-encoders method loses them at the end of the learning process. Table \ref{table:results} illustrates the examples of classification results in the first experiment. For each, we report the image patch, corresponding prediction result and the ground truth. Those normal tubules we identify as tubules with vacuoles and the other way around are extremely difficult to determine whether they are normal tubules or tubules with vacuoles. They are in essence ambiguous in terms of vacuoles. Even for those biologists with expertise in this area, they probably would not always reach a consistent classification result.  \\

\vspace*{-3.5em}
\begin{table}[!h]
\centering
\caption{Examples of our models classification results compared to the ground truth. The normal tubules that we identified with vacuoles are arguable since most appear to exhibit vacuole-like features. The tubules with vacuoles that we classified as normal tended to have few vacuoles and appeared out of focus, thus, leading to ambiguity. \\}
\begin{tabular}{c | c c c c | c c c}
\toprule[0.04cm]
&	\multicolumn{7}{c}{True class}\\
\hline
&	&	\multicolumn{3}{c |}{Normal}	&	\multicolumn{3}{c}{Vacuoles} \\
\multirow{2}{*}{Prediction}	&	Normal	&	\includegraphics[width=0.09\textwidth]{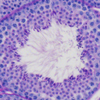} 	&	\includegraphics[width=0.09\textwidth]{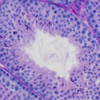}	&	\includegraphics[width=0.09\textwidth]{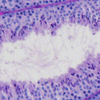}	&	\includegraphics[width=0.09\textwidth]{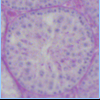}	&	\includegraphics[width=0.09\textwidth]{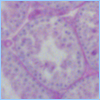}	&	\includegraphics[width=0.09\textwidth]{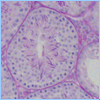}\\
&	Vacuoles	&	\includegraphics[width=0.09\textwidth]{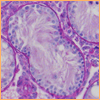} 	&	\includegraphics[width=0.09\textwidth]{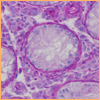}	&	\includegraphics[width=0.09\textwidth]{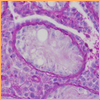}	&	\includegraphics[width=0.09\textwidth]{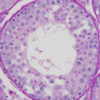}	&	\includegraphics[width=0.09\textwidth]{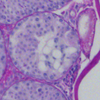}	&	\includegraphics[width=0.09\textwidth]{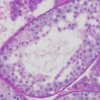}\\
\bottomrule[0.04cm]
\end{tabular}
\label{table:results}
\end{table}
\vspace*{-5em}
\begin{table}[!h]
\centering
\caption{Examples of reconstructed image generated from the deconvolutional networks using the auto-encoder features. Notice how the reconstructed image becomes blurrier at higher layer because max-pooling and convolution tend to decrease the resolution and smooth the images at higher layers.\\}
\begin{tabular}{c c c c c c}
\toprule[0.04cm]
	&	\parbox{1cm}{\centering Original \\Image}	&	\parbox{1cm}{\centering Norm \\ Image}	&	Layer 1		&	Layer 2	&	Layer 3\\
\hline
Normal			&	\includegraphics[width=0.09\textwidth]{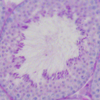} &	\includegraphics[width=0.09\textwidth]{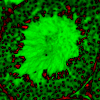} &	\includegraphics[width=0.09\textwidth]{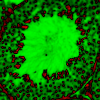} &	\includegraphics[width=0.09\textwidth]{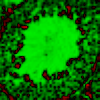}	&	\includegraphics[width=0.09\textwidth]{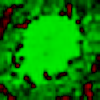} \\
\vspace*{0.01em}
Vacuoles			&	\includegraphics[width=0.09\textwidth]{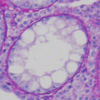} &	\includegraphics[width=0.09\textwidth]{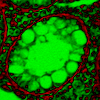} &	\includegraphics[width=0.09\textwidth]{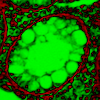} &	\includegraphics[width=0.09\textwidth]{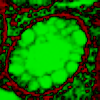}	&	\includegraphics[width=0.09\textwidth]{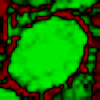} \\
\bottomrule[0.04cm]
\end{tabular}
\label{table:recon}
\end{table}

\vspace*{-2em}
\noindent{}\textbf{Image Reconstruction with Deconvolutional Networks:}
Table \ref{table:recon} shows the results of image reconstructions via deconvolutional networks. We normalized the images before learning the features to alleviate the differences of color distribution stemming from hematoxylin stained. We highlight the values below zero in red and above in green. Because of max-pooling following by the convolutional layer, the resolution of the images decreases as the layer goes up. In addition to the max-pooling layer, convolution operation smooths the image while applying the filter over the whole image. Therefore, we can observe that the reconstructed image becomes blurrier at the higher layer as expected. The reconstructed image generated via deconvolutional networks takes the weight after fine-tuning stage and propagates the results through the bottom layer. We adopt the tuned weights to reconstruct the image, and hence, the image reflects the classification features learned from the model. 

\vspace*{-1em}
\section{Conclusion}
\vspace*{-0.5em}
We present a novel deep learning architecture for features learning and automated tubule classification for testis histology images. Based on stacked convolutional auto-encoders, we insert a hyperlayer, collecting features from every convolutional layer below, to increase the perception of the classifier. The approach outperforms canonical deep networks and shows an improvement over the existing methods. Plus, we construct an extension of the architecture to see how well the features learned by reconstructing the images from top to bottom. The work helps provide a better understanding of the prediction produced by the automated classifier.

\end{document}